**When Bots Join the Team: Bot Adoption and the Institutional Fabric of Open-Source Software Projects**

Yongren Shi (yongren@arizona.edu)

Wenyi Gong (wenyig@arizona.edu)

University of Arizona

Manuscript: July 15th, 2026

## Abstract

AI agents are joining human teams, raising a basic question: when an automated agent becomes a regular participant, does group organization strengthen or weaken? We study this question in open-source software, where bots open pull requests, review code, and merge changes alongside people, leaving a public record of every interaction. Treating bots as participants rather than tools, we examine 2,991 GitHub projects for two years before and after each adopted its first bot. We measure three capabilities that institutional theory links to durable coordination—repeated engagement, social memory, and role differentiation—and two outcomes: conflict cascades and output distinctiveness. Bot adoption is followed by more repeated collaboration, greater recognition of specific bots in discussion, fewer conflict cascades, and more distinctive outputs. These changes cluster around adoption rather than accumulating gradually. Because we lack an untreated comparison group, we interpret the results as precisely timed associations, not causal effects. Two patterns are difficult for alternative explanations to account for: capabilities predict outcomes according to their function—coordination versus differentiation—rather than whether humans or bots provide them, and human-side capabilities account for the bot–conflict association but not the bot–distinctiveness association. The findings are consistent with a specific interpretation: predictable, rule-based agents can become part of a community's social infrastructure. The bot is the occasion; social organization is the mechanism.

## 1. Introduction

Artificial-intelligence (AI) agents — systems capable of perceiving their environment, making decisions, and taking autonomous action — are increasingly embedded in human collaborative work. AI agents now propose code changes in open-source software, assist physicians in clinical diagnosis, and contribute to experimental design in scientific research (Rahwan et al. 2019), and autonomous coding agents now author pull requests across hundreds of thousands of GitHub repositories (Hassan et al. 2025; AIDev 2026). Sociologists have called for a research program adequate to this shift: a group-level science of how humans and machines jointly produce social outcomes, beyond the human–machine dyad (Cerulo 2009; Tsvetkova et al. 2024). The empirical question is no longer whether automated agents will join human teams but what their presence coincides with in the organizational fabric that makes coordination possible.

This question presents a puzzle for sociological theory. Human groups coordinate not merely through technical efficiency but through *social institutions* — durable, self-reproducing patterns of interaction that align behavior, contain conflict, and enable collective innovation (Berger and Luckmann 1966; Giddens 1984; North 1990; Jepperson 1991). Institutions are not rules imposed from above; they are patterns that reproduce themselves through the capabilities of their participants. Contributors who remember who reviewed what, who keep encountering the same collaborators month after month, and who differentiate into distinct evaluative roles sustain an institutional pattern that prevents conflict and enables non-redundant exploration of the design space. The pattern persists not because it is mandated but because participants' capabilities reproduce it.

AI agents, as currently constituted, possess these capabilities only partially. They lack the reliable theory of mind that enables mutual anticipation. They cannot be sanctioned in the way humans can. Their memory is a product of configuration files rather than lived experience. Yet they are present, reliably and predictably, in ways that may alter the institutional fabric despite these limitations. When bots consistently participate in pull requests, are treated as recognizable actors, and handle specific domains of work, they become embedded in the project's interaction structure in ways that resemble human contributors. Whether such traces matter — whether predictable bot presence coincides with a reinforced institutional fabric or merely adds noise — is the question we take up.

Two competing intuitions frame the empirical stakes. The *displacement* intuition holds that automating routine work removes the interaction through which institutions form. If a bot handles dependency updates, for example, humans no longer need to coordinate around that work, and the social ties that would have formed through such coordination never develop. This intuition has empirical support: studies of bot adoption in open-source projects find no consistent gains in collaboration metrics afterward (Wessel et al. 2018), developers often experience bot activity as disruptive noise (Wessel et al. 2021), and the most rigorous causal study to date finds that code-review bots *reduce* human communication on pull requests even as they shorten merge times (Wessel et al. 2022). On this account, bot adoption should coincide with a weakened institutional fabric: declining repeated engagement, eroding social memory, and greater fragility under turnover. The *infrastructure* intuition holds the opposite. A bot's reliable presence adds a stable, named, repeatedly encountered participant to the project's shared record of interaction. Contributors form expectations around the bot's behavior, reference it by name in discussions, and differentiate their own roles in response. Experimental work supports this intuition as well: simple automated agents placed in human groups can improve collective coordination (Shirado and Christakis 2017), and a machine teammate's behavior can reshape how the humans in a team interact with one another (Traeger et al. 2020). On this account, bot adoption should coincide with a reinforced institutional fabric, strengthening the capabilities through which coordination and innovation are sustained.

This paper develops and tests a single claim: automated agents can become part of a community's institutional infrastructure because their participation alters the community's shared record of interaction — the publicly observable trace of who did what, toward whom, and with what consequences — from which institutions are reproduced. This record carries behavioral indicators of institutional capabilities, which in turn predict collective outcomes. Each link in this chain yields testable implications, which we examine using 2,991 open-source projects observed across bot adoption.

Three scope conditions define our claims. First, our measures are behavioral indicators, not institutions themselves. Contributor overlap, bot-referencing language, and action-mix entropy are narrow observable traces of broad constructs, so our findings should be read as behavioral signatures consistent with institutional reinforcement rather than as institutional reproduction observed in its entirety. Second, we study the coordinative dimension of institutions, not their roles in allocating power, conferring legitimacy, or constructing identities (DiMaggio and Powell 1983; Meyer and Jepperson 2000). Third, our analysis concerns rule-based, predictable, domain-specific, non-conversational bots in open-source communities; implications for more capable AI agents remain conjectural.

The same caution applies to causal interpretation. Our within-project event-study design, centered on each project's bot-adoption month, lacks an untreated comparison group, and adoption is plausibly endogenous to project conditions. We therefore interpret the results as precisely timed associations: bot adoption coincides with shifts in the predicted directions across the capability indicators and outcomes. We state the design's inferential limits, evaluate alternative explanations — direct automation, project maturation, platform change, and selection on transitory shocks — and identify stronger identification strategies for future work. The institutional interpretation rests not on any single estimate but on the convergence of evidence: the temporal alignment of changes with adoption, the specificity linking capability indicators to predicted outcomes, and the asymmetric attenuation of bot-intensity associations.

Prior work on bots in open-source software has measured their association with output — activity volume, merge and rejection times, contributor counts — with mixed results (Wessel et al. 2018, 2022; Wessel and Steinmacher 2020; Rebatchi et al. 2024). Adjacent literatures frame automated systems as objects of reliance and trust in experimental teams (O'Neill et al. 2022) or as instruments of managerial control over workers (Kellogg, Valentine, and Christin 2020). We ask a different question, one that requires a system-level view (Anthony, Bechky, and Fayard 2023): what bot adoption coincides with in the social organization that sustains the work. Communication volume can fall while relational structure strengthens; throughput can remain unchanged while the social organization develops. The constructs differ, and so, we will show, do the answers.

## 2. Theoretic Framework: Bots, the Shared Record, and Institutional Reproduction

The framework is a chain with three links: bots enter the shared record of interaction; the record carries behavioral indicators of institutional capabilities; and the capabilities predict collective outcomes. Sections 2.1 through 2.3 develop the links in turn; Section 2.4 states the hypotheses.

### 2.1 AI Agents as Participants in Collaborative Production

By an AI agent we mean a system that perceives its environment, makes decisions, and takes autonomous action within a social setting. A bot in an open-source software project meets each criterion: it perceives the repository's event stream, evaluates it against a configured policy, and acts — opening pull requests, posting reviews, merging changes, closing inactive discussions — within the same interactional space that human participants occupy. The actions are taken under a name, leave timestamped traces, are addressable by other participants, and trigger responses from the human members of the community.

Two features distinguish an agent from a tool (Lebeuf, Storey, and Zagalsky 2018). First, an agent operates visibly and autonomously in the project's interaction space, rather than invisibly within a single user's workflow. The distinction matters: invisible AI assistants have been found to raise coordination costs in projects even as they raise individual output (Song, Agarwal, and Wen 2024), which is consistent with our premise that participation in the shared public record, not automation itself, carries the organizational consequences we study. Second, an agent has addressable identity: a stable name that other participants use and reference in subsequent interactions. Addressable identity makes agents part of the community rather than features of the technological environment. The distinction has observable behavioral consequences: experimental research shows that people apply social norms and expectations to machines that display even minimal social cues, such as a name, turn-taking, or consistent behavior (Nass and Moon 2000). An identifiable bot provides these cues continuously through the community's public record of interaction.

We emphasize at the outset what kind of agents these are: rule-based, template-driven, domain-specific, and non-conversational. They are the least cognitively sophisticated agents that qualify as participants under the definition above, which is what makes them theoretically useful. If participants this simple leave traces that matter institutionally, the mechanism cannot depend on rich cognition.

### 2.2 Institutional Capabilities for Coordination

Institutions perform many functions: they allocate power, define legitimacy, shape identities, and constitute the actors entitled to act (DiMaggio and Powell 1983; Meyer and Rowan 1977; Meyer and Jepperson 2000). This paper concerns one dimension of institutional life: the coordinative dimension — the reproduction of the recurrent interaction patterns through which a collective aligns behavior, contains conflict, and sustains non-redundant exploration (Berger and Luckmann 1966; Giddens 1984; Jepperson 1991; Zucker 1977). We set aside questions of power, legitimacy, and identity not because they are unimportant but because our archival measures cannot reach them. Within this dimension, our approach is microsociological: we follow the strand of institutionalism that locates reproduction in situated interaction, in the participants who inhabit institutions and perform their patterns (Powell and Colyvas 2008; Hallett and Ventresca 2006; Feldman and Pentland 2003), and in self-governing production communities, where institutions demonstrably emerge from and are policed through member interaction rather than external mandate (Ostrom 1990; O'Mahony and Ferraro 2007).

Institutionalization, as Zucker (1977) showed, is a variable rather than a binary. This carries a methodological implication: rather than asking whether a project has institutions, we ask how strong the capabilities sustaining its coordinative patterns are, and whether capability strength predicts the outcomes those patterns are supposed to produce. Three capabilities, available in principle to any participant, are central.

*Social memory.* Coordination requires participants to retain knowledge of past interactions — not only that events occurred, but who did what to whom. Sociologists have long located such memory beyond the individual, in collective mnemonic practices (Halbwachs 1992 [1925]; Olick and Robbins 1998) and in the external artifacts and records that carry cognitive work in real settings (Hutchins 1995). Cognitive accounts instead emphasize memory distributed across individuals and their knowledge of one another (Wegner 1987; Lewis 2003). We do not adjudicate between these perspectives. We focus on the observable record through which memory becomes available for coordination, and through which low-cognition agents can participate. Although bots lack episodic memory, their repeated contributions leave traces in the shared record from which participants form expectations (Esposito 2022).

*Repeated engagement.* Coordination requires sustained interaction among participants. The expectation of future contact encourages investment in coordination and builds the familiarity underlying trust (Axelrod 1984; Kollock 1994; Uzzi 1997), and repeated focused interaction generates the solidarity and shared expectations on which cooperation rests (Collins 2004). In software teams, experience working together predicts coordination beyond what individual experience predicts (Reagans, Argote, and Brooks 2005), and familiarity reduces reliance on explicit coordination in distributed work (Espinosa et al. 2007). Our framework assigns repeated engagement one specific function: preventing conflict cascades, in which contributors repeatedly overwrite one another's work. Such cascades mark failures of coordination in open-source projects, with direct parallels in Wikipedia revert conflicts (Yasseri et al. 2012; Halfaker et al. 2013).

*Role differentiation.* Coordinative patterns are sustained not only through repeated interaction but also through specialization among participants (Stinchcombe 1965; Padgett and Powell 2012). Specialization is common in open-source communities (von Krogh, Spaeth, and Lakhani 2003). Role differentiation serves two functions: it divides tasks across domains, and it expands the range of perspectives available for collective exploration (March 1991; Page 2007).

Together, these three capabilities are expressed in the project's *shared record of interaction* — the timestamped, addressable, publicly visible trace of who acted, toward whom, when, and with what consequences. The record is where structure, in Sewell's (1992) sense, is reproduced through observable practice. It is not an alternative to cognitive accounts of coordination, and we do not claim that records replace minds. It is the observable layer through which institutional reproduction leaves measurable traces, situated between cognitive accounts that place memory in minds (Wegner 1987; Lewis 2003) and structural accounts that place coordination in network topology (Burt 1992). Focusing on this layer allows us to study coordination across both human and low-cognition participants: whether a bot's contributions matter because human actors interpret them or because the record itself structures action, institutional participation must enter through observable traces.

### 2.3 Bots as Contributors to the Shared Record

If coordination depends on the shared record of interaction, then even an agent with limited cognition can contribute, provided its actions leave publicly observable, interpretable traces. A bot that predictably addresses specific contributors becomes part of this record, especially when others respond to or reference its actions in subsequent interactions. At that point it functions less as a tool than as a participant in coordination (Seeber et al. 2020).

We do not argue that bots and humans are interchangeable. Bots lack theory of mind, cannot be sanctioned in the conventional sense, and rely on programmed rules rather than experience (Premack and Woodruff 1978; Mead 1934; Heckathorn 1990; cf. Strachan et al. 2024). Our claim is one of functional equivalence at the level of the shared record: if a bot's recurring traces support the same expectation formation as those of a human contributor, they can reinforce coordination regardless of the mechanism inside the agent. This view is related to theories that attribute social significance to nonhuman actors (Latour 1992; Nass and Moon 2000; Cerulo 2009), but it advances a narrower and testable claim about community-level organization.

Two features make bots consequential contributors to the shared record. First, they are unusually stable participants. Unlike human contributors, whose participation varies over time, bots recur with mechanical regularity, providing a reliable basis for repeated engagement. Second, bots are addressable: they have persistent identities, are named and referenced by contributors, and become associated with specific domains of work. Addressability gives participants a stable object for forming expectations about future behavior.

These features motivate two bot-side capability indicators, defined by the traces bots leave rather than by their internal cognition. Bot-to-human addressing captures the extent to which bots direct actions toward specific contributors rather than broadcasting; addressing assigns responsibility and should therefore reduce conflict. Inter-bot interaction captures the extent to which bots act on one another's outputs, composing specialized workflows that expand the project's repertoire of coordinated action; chaining should therefore increase distinctiveness. Our structural prediction is that these indicators align with the functions their traces perform — coordination or differentiation — rather than with whether the contributor is human or automated.

### 2.4 Hypotheses

The framework yields five testable hypotheses.

The first concerns the direction of change at adoption, where the two accounts make opposite predictions. The displacement account holds that automation removes occasions for interaction, so the indicators of repeated engagement and social memory should decline after adoption. The infrastructure account holds that a stable, addressable participant provides an anchor for repeated encounters and a recognizable object of memory, so both indicators should rise.

**H1 (Capability reinforcement).** *Bot adoption is followed by increases in the indicators of repeated engagement and social memory in the post-adoption window relative to the pre-adoption window.*
If the indicators reflect real changes in coordination, the outcomes that coordination governs should change as well, and on the same schedule. Because the proposed mechanism begins when the new participant enters the record, the changes should be concentrated at the adoption month. Gradual change would be equally consistent with project maturation and would therefore be less informative.

**H2 (System-level outcomes).** *Bot adoption is followed by a decline in conflict cascades and a rise in output distinctiveness, with the shifts concentrated at the adoption month.*
The framework also specifies which capability should predict which outcome, including where no association is expected. Repeated engagement supports anticipatory coordination: contributors who repeatedly encounter one another learn each other's domains of work and avoid conflicting changes. It supplies no new perspectives, however, so it should not predict distinctiveness. Role differentiation multiplies the evaluative perspectives applied to a shared design space, so differentiated projects should produce more distinctive output; because specialization also separates task domains, it is the one capability expected to predict both outcomes. The claim concerns evaluative differentiation specifically, so we test both a narrow review-state measure and a broad eight-action measure (Section 5.3). The bot-side indicators follow the same logic: bot-to-human addressing should predict less conflict, and inter-bot interaction should predict greater distinctiveness (Section 7.4).

**H3 (Repeated engagement and conflict cascades).** *Projects with stronger repeated-engagement indicators exhibit fewer conflict cascades.*
**H4 (Role differentiation and output distinctiveness).** *Projects whose contributors differentiate into distinct evaluative roles exhibit greater within-subdomain output distinctiveness.*

The final implication distinguishes the institutional account from the alternative that bots are simply efficient tools. If bots relate to conflict through the social organization of the project, then controlling for the capability indicators should reduce the bot-intensity association toward zero. If bots reduce conflict directly, the association should remain after those controls.

**H5 (Attenuation).** *The association between bot intensity and conflict cascades attenuates toward zero when the repeated-engagement and role-differentiation indicators are controlled.*

## 3. Research Setting: Open-Source Software as a Site for Institutional Analysis

Open-source software (OSS) projects combine features that make them suitable for studying coordinative dynamics at scale. Every interaction is publicly recorded, and contributors read this visible record to infer what others are doing and coordinate accordingly (Dabbish et al. 2012). Most projects follow pull-based development, in which contributors propose changes that others review and integrate. Work is ideally organized so that each contribution can be added independently, without modifying the parts of the code that others are changing (Howison and Crowston 2014). Governance is largely endogenous: authority structures emerge from member interaction (O'Mahony and Ferraro 2007), and contributors join through recognizable scripts and settle into specialized niches (von Krogh, Spaeth, and Lakhani 2003). Maintenance burden is chronic and concentrated, which is why projects turn to automation (Eghbal 2020) and why adoption timing must be treated as endogenous.

Contributors, including bots, are identifiable by stable usernames. Bots are not new to peer production: on Wikipedia they have long acted as visible participants in community work and governance (Geiger 2011, 2014). The adoption of a bot is observable as a discrete event. Together these features provide what institutional analysis rarely has: a complete, timestamped behavioral record of a community before and after a nonhuman participant joins it.

## 4. Data

We constructed a sample of 3,000 public GitHub projects that adopted at least one bot between 2016 and 2025, using stratified random sampling from a candidate pool of 657,115 (repository, bot) pairs identified across the GitHub Archive 2016–2025 monthly tables. Two choices governed the sampling.

First, we stratified the corpus by bot category, drawing 78 percent of cases from task-automating bots, 20 percent from governance-participating bots, and 2 percent from communication bots — proportions that reflect the empirical distribution of the three bot types in the candidate pool (Table 1). Second, we capped the share of the sample attributable to any single bot at 40 percent. Dependabot alone accounts for a large share of bot adoptions on GitHub; a higher cap would have made the corpus a study of that single bot rather than of bot adoption in general.

The three categories differ in the kind of work they automate. Task-automating bots (78 percent) perform routine technical work — updating dependencies, formatting code, running tests — and produce proposals that humans evaluate. Governance-participating bots (20 percent) perform decisional actions on contributions, including automatic merging of pull requests that meet specified criteria. Communication bots (2 percent) greet new contributors, manage labels, and close stale discussions. In the analyses below, bot category enters as a fixed effect; category-specific heterogeneity is theoretically expected and deferred to future work.

| Bot category | n | % | Most common bots |
|---|---|---|---|
| Task-automating | 2,332 | 78.0 | dependabot[bot] (n = 1,200); renovate[bot] (n = 258) |
| Governance-participating | 599 | 20.0 | mergify[bot] (n = 379); codecov[bot] (n = 231) |
| Communication | 60 | 2.0 | — |

*Table 1. Sample composition (N = 2,991 repositories, 2016–2025).*

Of the 3,000 projects, nine were excluded because the repositories had been deleted, made private, or returned malformed data during export. For the remaining 2,991 projects we collected 49 monthly snapshots: 24 months before adoption (T−24 through T−1), the adoption month (T), and 24 months after (T+1 through T+24). The adoption month (T) is defined as the first month in which a bot's activity becomes sustained— active in that month plus at least two of the following six months — excluding one-time bot experiments that never became part of a project's routine. For each project-month we collected all comments on pull requests and issues with author and timestamp; all review actions with reviewer identity; the monthly contributor-interaction graph; the set of active human contributors; and a complete record of overwrite events derived from the project's version-control history. The top five primary languages are TypeScript (17.9%), JavaScript (15.9%), Python (14.1%), Go (6.7%), and Java (5.7%); fifteen languages account for 81% of the sample.

The 2,991 projects adopted bots across the full 2016–2025 span. The modal adoption year is 2020 (n = 622); the 2022–2024 group (combined n = 1,100) reflects continued growth of bot-mediated workflows. The dispersion of adoption dates across a decade matters for identification: no single calendar-time event can align with every project's adoption month.

## 5. Measures

Each construct below is a broad sociological category; each measure is a narrow behavioral indicator of it. We use the construct names as shorthand, but every claim should be read at the indicator level.

### 5.1 Repeated Engagement

We operationalize repeated engagement with two complementary set-overlap statistics computed over active human contributors (bot accounts are excluded, so the indicator cannot increase mechanically with bot activity). For each month $t$, we compute the Jaccard similarity between the set of contributors active in month $t$ and the set active across months $t-1$ through $t-3$, capturing whether the same individuals appear across short windows. We also compute retention as the fraction of contributors active in month $t$ who remain active by month $t+3$, capturing persistence over longer horizons. The composite averages the two across the relevant 12-month window; we compute pre- and post-adoption versions for the H1 test. Two components are used because repeated engagement has two aspects — short-term stability and longer-term persistence — and examining them separately lets us verify that they move together, ruling out compositional artifacts.

### 5.2 Bot-Directed Social Memory

Social memory, as introduced earlier, is memory of who did what, registered in the record. Our indicator measures the part of this construct that bot adoption makes observable: whether humans refer to specific bots and to their past behavior. Because the indicator is bot-referential, its pre-adoption baseline is necessarily low; there is little bot to remember before one arrives. The H1 test should therefore be read not as memory rising from nothing, but as evidence of whether contributors come to treat the new agent as a specific participant with a behavioral history rather than as undifferentiated automation. The pre-adoption baseline is not zero because some bot mentions precede sustained adoption: one-time bot experiments excluded by our adoption definition, bots in neighboring projects, and generic references to automation.

Two components make up the composite. Prior-reference rate counts, per post-T month, comment bodies matching a roughly 200-phrase regular-expression bank capturing temporal memory of bots — "again," "as before," "every time," "keeps doing this," and named-bot prefixes ("dependabot keeps," "renovate usually"). The measure is capped at min(1.0, n_match/5.0) per month to bound very long threads. Bot-name retention is the fraction, among all post-T comments mentioning bots or automation, that name a specific bot rather than "the bot" or "automation" generically. The composite averages the two components, each bounded in [0, 1], across post-T months. The texts capture the indicator directly as enacted in language: a contributor who writes "Dependabot is opening another duplicate PR again" is naming a specific agent and referencing its prior behavior.

### 5.3 Role Differentiation

GitHub contributors perform distinct action types: proposing changes (pull requests), merging, opening issues, commenting, and reviewing others' changes by approving, requesting revisions, or commenting. For each contributor active in the post-T window, we measure activity concentration as the entropy deficit of the action mix: $1 - H(action_mix)/\log(K)$, where H is Shannon entropy and K is the number of distinct action types performed. A contributor who performs only one action type scores 1.0; one who splits evenly among many scores 0.0. The project-level score averages across post-T contributors. The primary indicator restricts attention to the three review states — approved, changes_requested, commented — capturing evaluative differentiation: how contributors diverge in judging one another's work. Because the framework predicts evaluative differentiation rather than general division of labor, we report both the narrow review-state measure and the broad eight-action measure and compare them in Section 7.

### 5.4 Conflict Cascades

A cascade is a chain of overwrite events (minimum length two, within a 30-day window) in which each contributor modifies code lines recently touched by a different contributor. Detection uses line-level git blame, which attributes each line to the commit and author that last modified it. The project's cascade rate is the number of cascade chains divided by active-contributor-months (ACM), where ACM sums the months in which each human contributor was active. A minimum chain length of two distinguishes cascades from ordinary single-overwrite churn; the 30-day window matches the empirical median time-to-second-touch in typical OSS workflows; normalizing by ACM makes the rate comparable across project sizes. Line-level blame removes false positives from file-overlap detection: commits touching the same file but different lines are not counted as overwrites. The construct has direct peer-production precedents: revert chains are the behavioral signature of conflict on Wikipedia (Yasseri et al. 2012; Halfaker et al. 2013), and cascades are their version-control analogue.

### 5.5 Output Distinctiveness

Output distinctiveness captures the extent to which projects within the same subdomain pursue distinct approaches rather than converging on a common template. For each pair of projects in the same programming-language subdomain, we compute four pairwise distances: architectural distance (file/folder structure), library distance (dependency overlap), comment-style distance (TF-IDF on code comments), and review-pattern distance (action-mix distributions). Each project's composite distinctiveness score is its average distance from same-subdomain peers.

A project can be distinctive because it is idiosyncratic, poorly maintained, or stylistically unusual. Three considerations support the exploration interpretation. First, prior work treats within-category distinctiveness as a meaningful, measurable property of an organization's position among its peers (Zhao et al. 2017) and as the observable trace of non-redundant search in collective production (Uzzi and Spiro 2005), and sociologists have long observed that small groups develop distinctive local cultures through their own interaction (Fine 1979). Second, the framework makes a falsifiable prediction that the idiosyncrasy account does not: only the differentiation capabilities should predict distinctiveness (Section 7.4). Third, external validation remains the appropriate standard. Distinctiveness should correlate with external markers of valued novelty, such as forks, stars, downstream dependents, or novel dependency combinations; we identify this validation as a priority. Finally, one component of the measure (review-pattern distance) is partially coupled to bot adoption by construction, because bot actions enter the action-mix distributions.

### 5.6 Bot-Side Trace Indicators

The two bot-side indicators conceptualized in Section 2.3 are measured from the same event stream. Bot-to-human addressing captures the extent to which the project's bots direct actions at specific humans: the rate at which bot-authored comments and review actions @-mention or assign a specific human contributor, normalized by ACM. Inter-bot interaction captures the extent to which bots act on one another's outputs: the rate of bot-authored events in threads or pull requests initiated by a different bot, normalized by ACM.

### 5.7 Bot Intensity

We measure how pervasively bots participate in the post-T interaction space with three indicators: share of threads with bot (share of issue/PR threads with at least one bot comment), bot events per ACM, and bot–human pairs per ACM (distinct human × bot co-thread pairs per ACM). Table 2 reports descriptive statistics for all constructs before and after adoption.

| Construct | Pre n | Pre M | Pre SD | Pre Min | Pre Max | Post n | Post M | Post SD | Post Min | Post Max |
|---|---|---|---|---|---|---|---|---|---|---|
| Repeated engagement | 2,990 | 0.224 | 0.241 | 0.000 | 1.000 | 2,990 | 0.400 | 0.268 | 0.000 | 1.000 |
| Bot-directed social memory | 2,990 | 0.016 | 0.056 | 0.000 | 0.740 | 2,990 | 0.062 | 0.104 | 0.000 | 0.658 |
| Role differentiation | 2,990 | 0.039 | 0.114 | 0.000 | 0.970 | 2,990 | 0.106 | 0.189 | 0.000 | 1.000 |
| Cascade rate | 2,383 | 0.649 | 2.044 | 0.000 | 36.500 | 2,868 | 0.249 | 0.516 | 0.000 | 10.667 |
| Output distinctiveness | 2,990 | 0.283 | 0.392 | 0.000 | 1.896 | 2,990 | 0.499 | 0.480 | 0.000 | 1.923 |
| Share of threads with bot involvement | 2,990 | 0.059 | 0.160 | 0.000 | 1.000 | 2,990 | 0.276 | 0.299 | 0.000 | 1.000 |
| Share of PRs with bot review | 2,990 | 0.003 | 0.032 | 0.000 | 0.733 | 2,990 | 0.038 | 0.152 | 0.000 | 1.000 |
| Bot events per ACM | 2,383 | 4.192 | 70.090 | 0.000 | 3,348 | 2,868 | 14.489 | 66.134 | 0.000 | 2,696 |

| Construct | Pre n | Pre M | Pre SD | Pre Min | Pre Max | Post n | Post M | Post SD | Post Min | Post Max |
| --- | --- | --- | --- | --- | --- | --- | --- | --- | --- | --- |
| Bot–human repeat pairs | 2,383 | 0.081 | 0.469 | 0.000 | 14.364 | 2,868 | 0.291 | 1.292 | 0.000 | 23.833 |

*Table 2. Descriptive statistics, pre-adoption (et = −12 to −1) and post-adoption (et = +1 to +12).*

## 6. Analytical Strategy

### 6.1 Paired Pre/Post Tests

To test H1 and the pre/post component of H2, for each project we compare the 12-month pre-adoption window (T−12 to T−1) against the 12-month post-adoption window (T+1 to T+12) on each construct. Paired Wilcoxon signed-rank tests provide distribution-free significance; Cohen's *d*, computed on paired differences, provides standardized effect sizes.

### 6.2 Event-Study Estimates, and What They Can and Cannot Establish

To assess whether outcomes shift abruptly at adoption or drift gradually, we estimate an event-study specification — a regression discontinuity in time (Hausman and Rapson 2018) — on the monthly panel spanning 24 months before to 24 months after adoption:

$$y_{it} = \alpha_i + \sum_{k \neq -1} \beta_k \cdot \mathbf{1}\{e_{it} = k\} + \varepsilon_{it}$$

where $y_{it}$ is the outcome for project *i* in month *t*; $e_{it}$ is event time (months from adoption); $\alpha_i$ is a project fixed effect absorbing all time-invariant differences across projects; and the month before adoption (k = −1) is the omitted reference, so each $\beta_k$ measures the average difference at event time *k* relative to that baseline. Standard errors are clustered by project. For count outcomes we use log((y + 1)/(active humans + 1)), a size-normalized log-rate that handles zeros.

There is no untreated comparison group: every project in the sample adopts, and the counterfactual — what these projects would have looked like without adoption — is not directly observed. The project fixed effects remove stable between-project differences, and the event-time coefficients establish timing: whether change is concentrated at each project's own adoption month or distributed smoothly around it. Timing evidence constrains, but does not eliminate, confounding. It rules out explanations that predict smooth change (secular trends, gradual maturation) but not explanations that predict change coincident with adoption (adoption as a response to an acute shock; adoption bundled with other simultaneous reforms). The pre-adoption coefficients provide the standard diagnostic (Roth 2022): flat pre-trends support attributing a break at et = 0 to something that happened at adoption; a sloped pre-trend shifts the question to whether the break exceeds what extrapolation would predict.

One pre-trend shape deserves notice because it appears in our data. Cascade conflict rises into the adoption month. A rising pre-trend followed by decline is the characteristic signature of selection on a transitory shock — the "Ashenfelter dip" of program evaluation, in which units select into treatment at the peak of a temporary deviation and then revert regardless of treatment (Ashenfelter 1978; Heckman and Smith 1999). On that account, projects adopt bots when conflict rises sharply, and the post-adoption decline is regression to the mean. We evaluate this account against the estimates in Section 7.3 and systematically in Section 8.3, and we state the design's residual limits here: neither mean reversion nor adoption-coincident confounds can be conclusively excluded without a comparison group. The designs that would tighten identification — staggered difference-in-differences using later-treated projects as controls for earlier-treated ones (Callaway and Sant'Anna 2021; Sun and Abraham 2021; Goodman-Bacon 2021), matched non-adopter comparisons, placebo adoption dates, and synthetic controls — are identified in Section 8.5 as the priority next step.

### 6.3 Linking Capability Indicators to Outcomes

To test H3 and H4, we regress each outcome on the five capability indicators, bivariately (one indicator with controls) and jointly (all five simultaneously). Note the estimand shift: H1 and H2 concern within-project change at adoption; H3 and H4 concern cross-sectional variation — whether projects with stronger post-adoption indicators exhibit better outcomes, conditional on controls. The designs are complementary: the within-project results establish that adoption, indicators, and outcomes move together in time; the cross-sectional results test whether indicator strength, however arrived at, is associated with the outcomes the framework assigns to it. Let $i$ index projects, $C_{ik}$ denote indicator k ∈ {1,…,5}, and $X_i$ denote controls: log project size, log activity volume, and bot-category fixed effects.

For the cascade outcome, a negative-binomial GLM with $\log(ACM_i)$ offset:

$$\log(\mu_i) = \alpha + \sum_{k=1}^{5} \beta_k C_{ik} + \delta' X_i + \log(ACM_i)$$

For distinctiveness, OLS with HC1 heteroskedasticity-consistent standard errors:

$$D_i = \alpha + \sum_{k=1}^{5} \beta_k C_{ik} + \delta' X_i + \varepsilon_i$$

### 6.4 Bot Intensity and Attenuation

To test H5, we examine whether the association between bot intensity and each outcome remains after controlling for the capability indicators. The logic is attenuation: if bot intensity relates to cascades through the capabilities, then controlling for repeated engagement and role differentiation should reduce the bot-intensity coefficient toward zero; if the relation is direct, the coefficient should remain. This analysis has a known interpretive limit. A causal mediation reading requires sequential ignorability, meaning no unmeasured confounding of the mediator–outcome relationship, and archival data cannot establish that assumption (Imai, Keele, and Tingley 2010; VanderWeele 2015). We therefore interpret the results as attenuation patterns, showing which variables statistically account for which associations, and we describe them in those terms. The patterns are nonetheless informative as a comparative test: the efficient-tool account and the institutional account make opposite predictions about whether the capability controls absorb the bot-intensity association, so the data can show which prediction fails.

## 7. Results

### 7.1 Capability Indicators Rise After Adoption (H1)

Table 3 reports the paired pre/post comparisons. Both capability indicators increase substantially after adoption. Repeated engagement rises from a pre-adoption mean of 0.224 to a post-adoption mean of 0.400 ($\Delta = +0.176$, Cohen's $d = 0.69$, Wilcoxon $p < .001$). The social-memory indicator rises from 0.016 to 0.062 ($\Delta = +0.045$, $d = 0.54$, $p < .001$). The two components of each composite move together, ruling out compositional artifacts. In substantive terms, a typical project's contributor overlap across consecutive months rises from roughly 22 percent to roughly 40 percent. Because the contributor sets exclude bot accounts, this reflects human contributors re-encountering one another, not the bot inflating the overlap. In post-adoption discussion threads, roughly one in nine comments names a specific bot: contributors do not treat bots as an undifferentiated category but refer to them as individual, named agents.

| Construct | n | Pre mean | Post mean | Δ | Cohen's d | Wilcoxon p |
|---|---|---|---|---|---|---|
| Repeated engagement | 2,991 | 0.224 | 0.400 | +0.176 | +0.69 | < .001 |
| Bot-directed social memory (composite) | 2,991 | 0.016 | 0.062 | +0.045 | +0.54 | < .001 |
| — Prior-reference rate | 2,991 | 0.003 | 0.011 | +0.008 | +0.20 | < .001 |
| — Bot-name retention | 2,991 | 0.030 | 0.112 | +0.082 | +0.53 | < .001 |

*Table 3. Paired pre/post comparisons of capability indicators. The 12-month pre-adoption window (T−12 to T−1) is compared with the 12-month post-adoption window (T+1 to T+12). Effect sizes are Cohen's d for paired samples; p values from one-sided Wilcoxon signed-rank tests.*

The prior-reference component ($d = 0.20$) is smaller than bot-name retention ($d = 0.53$), consistent with explicitly temporal language ("the bot did this again") being rarer than naming the bot. Both move in the predicted direction. Recall the scope condition from Section 5.2: for the memory indicator, the informative result is not the rise from a low baseline, which is partly built into a bot-referential measure, but the form post-adoption memory takes — specific, named, and referenced in relation to past behavior. With that qualification explicit, H1 is supported.

### 7.2 System-Level Outcomes Shift at Adoption (H2)

Both outcomes shift in the predicted direction (Table 4). The cascade rate falls from a pre-adoption mean of 0.642 to a post-adoption mean of 0.229 ($\Delta = -0.413$, $d = -0.28$, $p < .001$); output distinctiveness rises from 0.283 to 0.499 ($\Delta = +0.215$, $d = +0.49$, $p < .001$). The cascade decline is not an artifact of the chain-detection procedure: the raw upstream overwrite count also falls (−5.57 events per ACM, $d = -0.22$), indicating a reduction in overwriting itself rather than a detection-threshold effect.

| Outcome | n | Pre mean | Post mean | Δ | Cohen's d | Wilcoxon p |
|---|---|---|---|---|---|---|
| Conflict cascade rate | 2,288 | 0.642 | 0.229 | −0.413 | −0.28 | < .001 |
| Output distinctiveness | 2,990 | 0.283 | 0.499 | +0.215 | +0.49 | < .001 |

*Table 4. Paired pre/post comparisons of system-level outcomes.*

### 7.3 The Shifts Are Concentrated at the Adoption Month

The event-study estimates show a sharp break at the adoption month rather than gradual accumulation (Table 5; Figures 1 and 2). At et = 0, the cascade log-rate falls by 0.204 ($SE$ = 0.015; 95% CI [−0.23, −0.17]; $z = -13.5$), implying a cascade rate of $\exp(-0.204) \approx 0.82$ relative to the et = −1 reference — an 18 percent reduction at the adoption month itself. The cascade rate continues to decline through et = 4 before partially reverting. Output distinctiveness jumps by +0.332 standard deviations at adoption ($SE$ = 0.012; 95% CI [+0.31, +0.35]; $z = +28.6$), and the two capability composites jump simultaneously (+0.149 and +0.112, respectively).

| Outcome | β at et = 0 | SE | 95% CI | z |
|---|---|---|---|---|
| Cascades (log-rate) | −0.204 | 0.015 | [−0.23, −0.17] | −13.5 |
| Output distinctiveness | +0.332 | 0.012 | [+0.31, +0.35] | +28.6 |

| Outcome | β at et = 0 | SE | 95% CI | z |
|---|---|---|---|---|
| Repeated engagement | +0.149 | 0.006 | [+0.14, +0.16] | +25.3 |
| Bot-directed social memory | +0.112 | 0.005 | [+0.10, +0.12] | +20.6 |
| Share of threads with bot | +0.269 | 0.008 | [+0.25, +0.29] | +32.2 |

*Table 5. Event-study estimates at the adoption month (et = 0), with et = −1 as reference. Project fixed effects; standard errors clustered at the project level (CR1).*

Figure 1 displays the pooled event-time trajectories by bot category for all four constructs. Three patterns are evident. First, the trajectories break at et = 0 — each project's own adoption month — across all three bot categories, despite adoption months being scattered over a decade of calendar time. Second, the capability indicators (Figure 1b, 1c) rise gradually before adoption and accelerate sharply at et = 0, with the jump at adoption exceeding any single pre-T step by roughly an order of magnitude (Figure 2 shows the coefficient-level view). Third, the cascade rate (Figure 1a) rises until the adoption month and then falls to a level below its earlier pre-adoption range, where it remains.

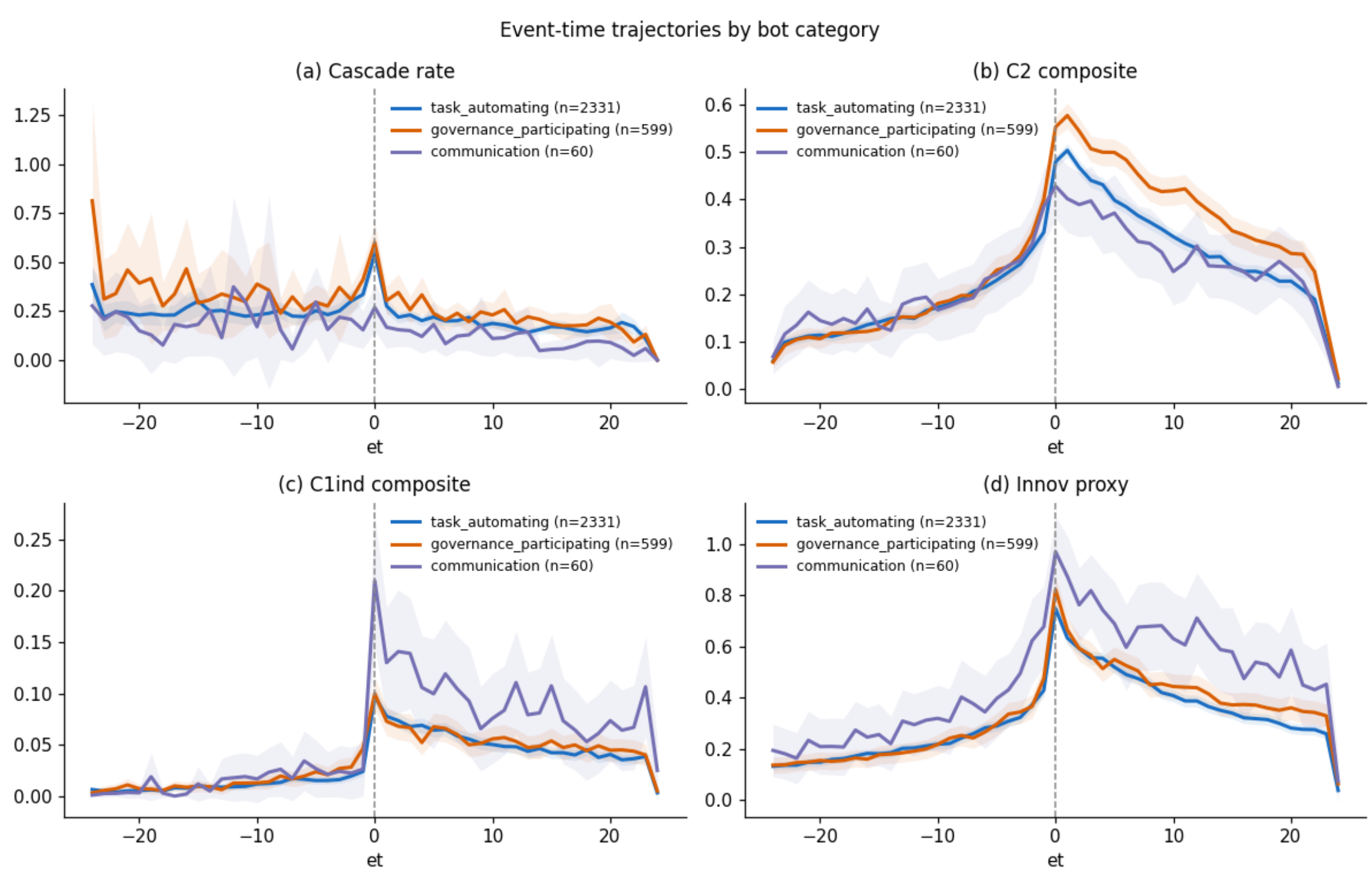

*Figure 1. Event-time trajectories by bot category, et = −24 to +24, with 95 percent confidence bands: (a) cascade rate, (b) repeated engagement composite, (c) bot-directed social memory composite, (d) output distinctiveness. Vertical dashed line marks the adoption month (et = 0).*

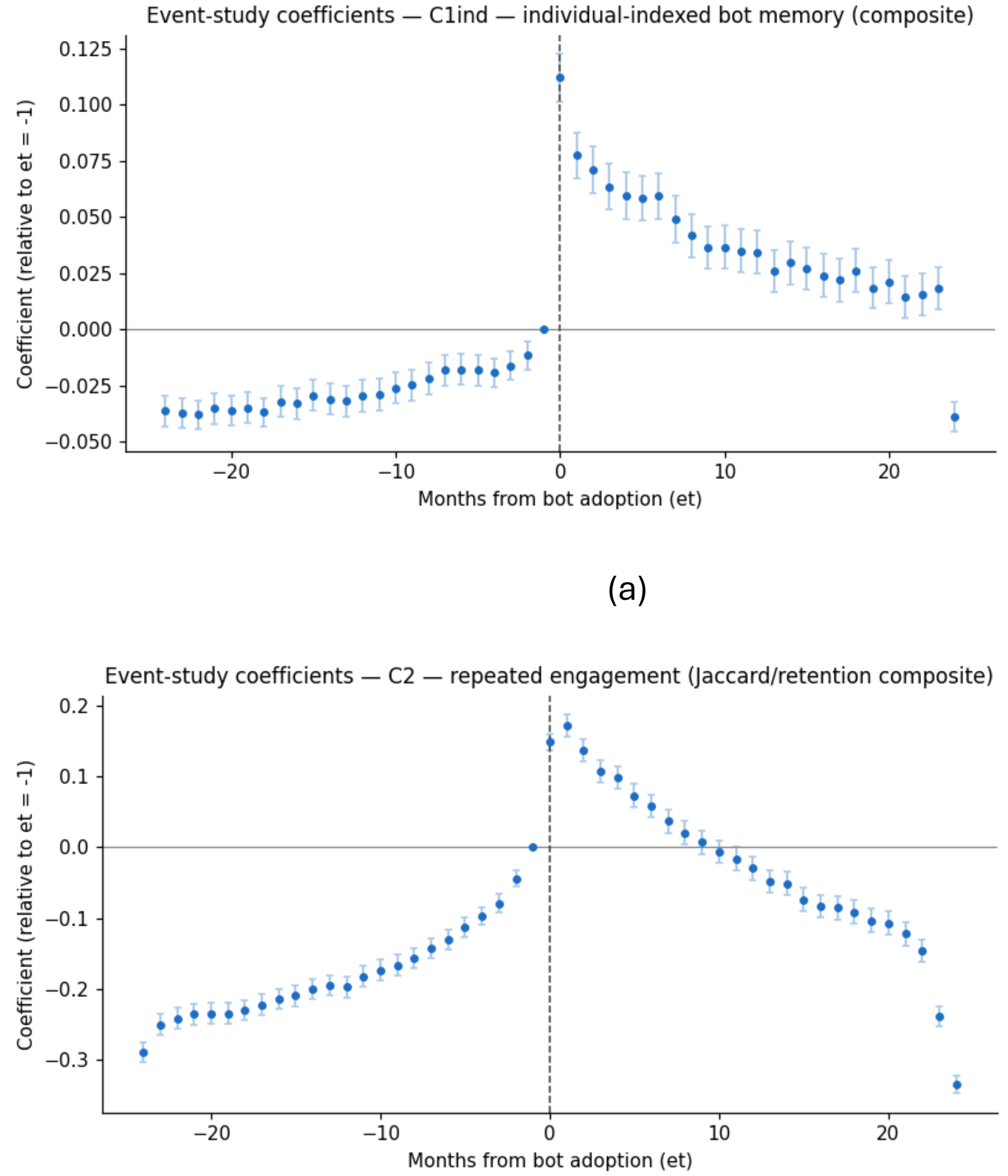

*Figure 2. Event-study coefficients (relative to et = −1, with 95 percent confidence intervals) for (a) bot-directed social memory and (b) repeated engagement. Both indicators drift upward before adoption and jump discontinuously at et = 0.*

The cascade trajectory is the most informative pattern for evaluating alternative explanations. Conflict was rising in the months before adoption; at et = 0 the trajectory reverses, with a coefficient of −0.20. A smooth secular trend, or a latent growth process that produced both cascades and adoption, would predict continued increase rather than a reversal at each project's own adoption month. The 95 percent confidence interval at et = 0 excludes the value implied by extrapolating the pre-adoption trend.

The mean-reversion alternative flagged in Section 6.2 can now be evaluated against the estimates, with the caveat that this is pattern comparison, not identification. If projects adopted at the peak of a transitory conflict shock, the post-adoption trajectory should revert toward the pre-shock baseline. Instead, the cascade rate falls below the level of the earlier pre-adoption months and remains there (Figure 1a), and the same month that reverses the conflict trajectory coincides with simultaneous jumps in repeated engagement, social memory, and distinctiveness (Figures 1b–d, 2) — constructs that a reverting conflict shock gives no reason to move, in any direction. These patterns are difficult to reconcile with pure mean reversion, but they do not exclude it, nor do they exclude adoption-coincident confounds; Section 8.3 addresses the full set of alternatives.

### 7.4 Capability Indicators and Outcomes: The Capability–Outcome Map (H3, H4)

Table 6 reports the joint five-indicator specifications (Section 6.3). The pattern is notable for its specificity. Three indicators are significantly associated with fewer conflict cascades — repeated engagement, bot-to-human addressing, and role differentiation — and three with greater output distinctiveness — role differentiation, social memory, and inter-bot interaction. Only role differentiation appears for both outcomes; every remaining cell is statistically indistinguishable from zero.

| Capability indicator | Conflict cascades | Output distinctiveness |
|---|---|---|
| Repeated engagement | **−0.411** ($p < .001$) | −0.037 ($p = 0.14$) |
| Bot-to-human addressing | **−0.740** ($p = 0.004$) | +0.019 ($p = 0.81$) |
| Role differentiation | **−0.303** ($p < .001$) | **+0.230** ($p < .001$) |
| Bot-directed social memory | +0.178 ($p = 0.43$) | **+0.192** ($p = 0.004$) |
| Inter-bot interaction | +0.056 ($p = 0.69$) | **+0.149** ($p < .001$) |
| n | 2,495 | 2,471 |

*Table 6. Regressions predicting conflict cascades (negative-binomial, ACM offset) and output distinctiveness (OLS, HC1). Controls: log size, log activity, bot-category fixed effects.*

These associations support H3 (repeated engagement and fewer cascades) and H4 (role differentiation and greater distinctiveness), and they match the structural prediction stated in Section 2.3: the indicators sort by functional mechanism, not actor type. Bot-to-human addressing behaves like repeated engagement — both associated with fewer cascades, neither with distinctiveness. Inter-bot interaction behaves like social memory — both associated with greater distinctiveness, neither with cascades. Coordination-type indicators cluster on the conflict outcome; differentiation-type indicators cluster on the distinctiveness outcome; role differentiation, which the framework assigns both functions, predicts both. This double dissociation is also the strongest evidence against a general project-health explanation, under which a single latent factor such as momentum, maturity, or professionalism would produce diffuse positive associations across all cells rather than the observed function-specific pattern.

### 7.5 Attenuation: Through the Indicators Versus Beyond Them (H5)

Table 7 reports the attenuation analysis. For cascades, the bot-intensity association attenuates essentially to zero once the capability indicators are controlled, moving from −0.354 ($p < .001$) to −0.015 ($p = 0.83$); repeated engagement (−0.294, $p < .001$) and role differentiation (−1.267, $p < .001$) carry the association. The pattern is consistent with the institutional interpretation — bot intensity relates to conflict through the capability indicators — and inconsistent with the efficient-tool account, under which the direct association should remain. H5 is supported in the attenuation sense defined in Section 6.4; we do not claim the pattern demonstrates mediation in the causal sense.

| Specification | Bot-intensity coefficient | p |
|---|---|---|
| **Conflict cascades** | | |
| Three-bot-intensity joint | −0.354 | < .001 |
| + Repeated engagement + role differentiation | −0.015 | 0.83 |
| **Output distinctiveness** | | |
| Three-bot-intensity joint | +0.162 | < .001 |
| + Repeated engagement + role differentiation | +0.156 | < .001 |

*Table 7. Attenuation analysis for bot intensity (share of threads with bot, primary indicator).*

For distinctiveness, the pattern is the opposite: the bot-intensity association remains essentially unchanged after capability controls (+0.162 to +0.156), and repeated engagement and role differentiation show null associations once bot intensity is in the model. Whatever links bots to distinctiveness is not captured by the measured capability indicators — a residual association we interpret in Section 8.2.

The asymmetry between the two panels of Table 7 deserves emphasis, because it is the single result least compatible with the alternative explanations. A generic confound — maturation, survival, platform change, project health — would have to explain why the bot-intensity association with one outcome is fully absorbed by two human-side behavioral indicators while its association with the other outcome is unchanged by the same controls. The institutional account predicts exactly this asymmetry: conflict containment operates through the human-side coordination capabilities that the bot's presence coincides with, while distinctiveness draws on operational novelty that the capability indicators do not measure.

## 8. Discussion

### 8.1 Behavioral Signatures of Institutional Reinforcement

Across 2,991 projects, bot adoption is followed by behavioral signatures consistent with a strengthened, not weakened, coordinative fabric. The capability indicators rise substantially (Table 3); conflict cascades fall and output distinctiveness rises (Table 4); the shifts concentrate at each project's adoption month (Table 5; Figures 1–2); the indicators predict outcomes according to their function (Table 6); and the bot-intensity association with conflict is fully accounted for by the human-side indicators while its association with distinctiveness is not (Table 7). Read at the indicator level, these results contradict the displacement account at each point where it made a prediction.

We interpret this pattern of results as evidence that highly predictable automated agents can acquire the behavioral signatures of institutional infrastructure: a relational substrate, embedded in practice, taken for granted until breakdown (Star and Ruhleder 1996; Star 1999). The claim is not that bots are universally beneficial. Automated agents can also degrade coordination: benign bots come into conflict with one another on Wikipedia (Tsvetkova et al. 2017), although much of this apparent conflict is routine coordination (Geiger and Halfaker 2017). Our claim is population-specific: in established OSS projects adopting their first bot, the dominant pattern is capability reinforcement. The proposed mechanisms are specific as well. Stability anchors repeated engagement; addressability anchors social memory. Both imply a testable pattern that the data confirm: projects whose bots leave traces of the kind human collaborators leave — addressing specific people by name, acting on other bots' work — show stronger institutional signatures than projects whose bots address no one in particular (Table 6).

### 8.2 Institution as Mechanism, Bot as Occasion

The attenuation asymmetry (Table 7) is the most direct comparison between the institutional account and the alternative that bots are efficient tools. If bots raised throughput without altering social organization, their association with cascade reduction should be direct and should remain after capability controls, and studies measuring only throughput would have captured their full effect. That such studies found little (Wessel et al. 2018), or found reduced communication (Wessel et al. 2022), is consistent with the effects being organizational rather than mechanical. In our data, the bot-intensity association with cascades is fully accounted for by two human-side indicators — the pattern expected if bots relate to conflict through changes in the social organization of work.

This reading places the study in a sociological lineage that predates AI. Barley (1986) showed that identical technology introduced into different organizations produced different structural outcomes, because the technology was an occasion for restructuring interaction rather than a cause acting alone; technology's organizational effects are realized through practice (Orlikowski 2000). Our results extend this insight to automated participants at population scale: the bot is the occasion; the social organization is the mechanism. A project could in principle achieve similar conflict reduction by any means that strengthens repeated engagement and role differentiation, such as stable recruitment or mentorship. What bots offer is reliability: they do not leave, reduce their involvement, or become unavailable.

For distinctiveness, the pattern differs. The bot-intensity association remains after capability controls (Table 7), so whatever connects bots to distinctive output is not captured by the measured indicators. One candidate, offered as interpretation rather than finding, is operational repertoire expansion: multi-bot configurations (Dependabot → codecov → Mergify chains; pre-commit-ci reformatting after linting bots) constitute novel operational compositions that differentiate a project's way of working from its peers. This is innovation through the recombination of available components (Baker and Nelson 2005; Padgett and Powell 2012). Competing candidates — signaling that attracts distinctive contributors, compositional differences in who joins — remain open (Section 8.6). The finding also has an unexpected implication for institutional theory: the canonical prediction under common institutional pressure is isomorphic convergence (DiMaggio and Powell 1983), yet the diffusion of a common technology here coincides with divergence, because what diffuses is not an output template but a combinable set of operational components.

### 8.3 Alternative Explanations

We consider four alternative explanations and evaluate each against three observed patterns: shifts concentrated at each project's own adoption month (Figures 1–2), the function-specific capability–outcome map (Table 6), and the asymmetric attenuation of bot-intensity associations (Table 7).

*Direct automation.* Conflict might fall because bots remove conflict-prone work: formatting becomes standardized, dependency updates arrive as bot pull requests, merge conflicts are resolved automatically. This account predicts a direct bot-intensity association with cascades that remains after capability controls, and it gives no reason for human contributor overlap to rise (Table 3) or for the capability–outcome map to be function-specific (Table 6). The attenuation result (Table 7) is the direct test, and the automation account fails it. The account retains a compositional version: automation may change which tasks humans perform, leaving work that is less prone to collision, and our data do not exclude this. A mechanical variant applies to distinctiveness: one of its four components (review-pattern distance) incorporates action-mix distributions that bot activity enters by construction (Section 5.5), so part of the jump at et = 0 could be definitional. The other three components are free of this coupling, and a component-level decomposition is the appropriate check.

*Maturation and survival.* Projects may stabilize as they age. Project fixed effects remove stable differences, and the cross-sectional models control for size and activity, but maturation is a within-project process that fixed effects do not absorb. Timing is the main evidence against it: maturation predicts smooth trajectories, not breaks at adoption months that vary across projects (Figure 1). A related concern is survivorship: the design requires 24 post-adoption months of observation, so the sample conditions on survival, and rising repeated engagement could partly reflect this. The indicators are defined even in low-activity months, which limits mechanical inflation, but matched comparisons are needed to resolve the concern.

*Platform and calendar change.* GitHub's infrastructure changed throughout the window: Dependabot became native in 2019, and the 2022–2024 cohort's post-adoption windows overlap the generative-AI period. Platform changes are calendar-time events, however, and adoption months are dispersed across a decade (Section 4), so no common calendar event can produce breaks aligned to each project's own adoption month. The remaining concern is cohort-specific — the large 2020 cohort could dominate pooled estimates — and cohort-stratified estimation is the check (Section 8.5).

*Selection on a transitory shock.* The most serious alternative: projects adopt bots at the peak of a conflict shock, and the decline is regression to the mean (Ashenfelter 1978; Heckman and Smith 1999). Two observations weigh against the pure version. The cascade rate falls below the earlier pre-adoption range and remains there (Figure 1a), and the adoption month coincides with simultaneous movements in three other constructs that a reverting conflict shock gives no reason to move. Against the broader version — adoption bundled with other simultaneous changes, such as governance reform or maintainer turnover — the design has no answer: anything that happens at adoption is absorbed into the adoption-month estimate. Only a comparison group can rule this out, and it is the first priority in Section 8.5.

Each alternative accounts for part of the evidence. None accounts for the combination of breaks timed to each project's own adoption month, a function-specific capability–outcome map, and asymmetric attenuation. The institutional interpretation is the most parsimonious account of that combination. It remains an interpretation, and our claims are stated accordingly.

### 8.4 The Capability–Outcome Map and Its Scope

The five-indicator matrix (Table 6) is the framework's strongest confirmation. Each indicator occupies a distinct cell: repeated engagement predicts fewer cascades; social memory predicts greater distinctiveness; role differentiation predicts both, because it separates task domains and multiplies evaluative perspectives. The bot-side indicators follow the function of their traces, not the nature of their author: addressing behaves like the coordination indicators, chaining like the differentiation indicators.

Within this population, what mattered was what a participant's traces do, not what the participant is. The scope of this conclusion must be stated precisely, because it is the paper's strongest claim and the most easily overextended. The agents studied are rule-based, template-driven, domain-specific, and non-conversational; the setting is open-source software, with its unusually complete public record; and the equivalence shown is at the level of trace function, not behavior — agent populations left to themselves behave in distinctly non-human ways (Goyal et al. 2026). Generalization to AI agents at large is a conjecture, and the framework specifies what it depends on: the institutional contribution of an automated participant should increase with the form of its traces — addressability, stability, chaining — rather than with its intelligence. More capable agents make this conjecture testable (Section 8.6).

The theoretical consequence is nonetheless real. Cognitive accounts of team coordination (Wegner 1987; Lewis 2003) would not predict that participants this simple could be associated with strengthened coordination. Our results show that they can be, through the same functional channels as human-side capabilities. This does not show that records replace minds; most plausibly, bot traces matter because human cognition processes them. It does show that the observable record is a level at which institutional analysis can proceed without settling the cognitive question (Esposito 2022; Knorr Cetina 1997), and at which participants without minds can be measured on the same terms as participants with them.

### 8.5 Limitations

Six limitations constrain interpretation, in descending order of consequence.

*First, identification.* The design lacks an untreated comparison group; adoption is plausibly endogenous to maintenance burden, contributor overload, governance change, and maturity; and anything bundled with adoption is absorbed into the adoption-month estimate. The priority analyses are staggered difference-in-differences with later-treated projects as controls (Callaway and Sant'Anna 2021; Sun and Abraham 2021; Goodman-Bacon 2021), matched non-adopter comparisons, placebo adoption dates, synthetic controls, and cohort-stratified estimation. Until then, our claims are stated as precisely timed associations.

*Second, measurement.* Every construct is measured by a narrow behavioral indicator. The gap between indicator and construct is largest for social memory (bot-referential by design; Section 5.2) and output distinctiveness (not innovation without external validation; Section 5.5). Validation against forks, stars, and downstream dependents is the concrete next step.

*Third, institutional scope.* Our evidence concerns the coordinative dimension of institutional reproduction. Institutions also allocate power, confer legitimacy, and construct identity, and bots plausibly matter there as well — governance-participating bots make decisions that allocate authority — but our indicators do not measure authority, legitimacy, or sanctioning. Mutual anticipation and norm enforcement likewise leave no archival trace and are reserved for experimental work. Claims about the institutional fabric should be read as claims about its coordinative dimension.

*Fourth, vendor determination.* The bot-side indicators reflect the adopted bot mix, not within-project learning: a project scores high on addressing largely because it adopted Mergify, whose templates mention reviewers by name. The trace-level interpretation is unaffected, but the causal locus is the adoption decision, not within-project adaptation.

*Fifth, corpus composition.* The corpus is English-language and concentrated in web development (TypeScript and JavaScript jointly 34 percent); generalization requires replication in other domains. The distinctiveness partition also fell back from dependency communities to language subdomains because dependency-overlap data were sparse; a finer partition would strengthen the H4 test.

*Sixth, calendar time.* The 2022–2024 cohort's post-adoption windows overlap the generative-AI period, and event-time pooling mixes months from different eras. The project-specific timing of the breaks limits, but does not eliminate, the concern; cohort-stratified estimates are the check.

### 8.6 Future Directions

Four directions follow. *Identification.* The staggered adoption in the sample already permits designs using later-treated projects as controls; executing them, with matched non-adopters and placebo dates, is the highest-priority next step. *Validation and extension of the outcomes.* Distinctiveness should be validated against external markers of valued novelty, and the institutional dimensions set aside here — authority, legitimacy, sanctioning — are measurable in principle from governance events such as maintainer appointments and permission changes. *Mechanisms for the distinctiveness residual.* The candidate mechanisms — repertoire expansion, signaling, compositional sorting — make different predictions about when distinctiveness should rise and in which projects, and can be separated with finer-grained designs. *More capable* *agents, and removal.* The bots studied here are the simplest agents that qualify as participants; the agents now arriving are not. Autonomous coding agents author pull requests at scale (Hassan et al. 2025; AIDev 2026), show partial theory-of-mind capacities (Strachan et al. 2024), and form populations whose own social dynamics are an emerging object of study (Holme and Tsvetkova 2025; Köster et al. 2022; Ashery, Aiello, and Baronchelli 2025). The framework's conjecture — institutional contribution increases with trace form, not intelligence — is directly testable on them. Removal provides the complementary test: infrastructure is most visible at breakdown (Star 1999). Bot deprecations and vendor shutdowns offer natural experiments; degradation of the indicators after removal would strengthen the causal interpretation, while persistence would suggest that bots initiate organization that then sustains itself.

## 9. Conclusion

We set out to adjudicate between two accounts of what happens when an automated agent becomes a regular participant in a human community: displacement, in which automation removes the interaction on which social organization depends, and infrastructure, in which a stable, addressable participant reinforces it. Across 2,991 open-source projects, the evidence favors infrastructure. Bot adoption is followed, at the month of adoption, by rising indicators of repeated engagement and social memory, falling conflict cascades, and rising output distinctiveness; the indicators predict outcomes according to the function of their traces rather than the actor supplying them; and the bot-intensity association with conflict is fully accounted for by human-side indicators while its association with distinctiveness is not. These are precisely timed associations rather than identified causal effects, and the alternatives — automation, maturation, platform change, selection — each remain in residual form. None of them explains the combination.

The contribution is one claim, now with large-scale evidence: automated agents can become part of a community's institutional infrastructure because their participation alters the shared record of interaction on which coordinative reproduction draws. The bot is the occasion; the social organization is the mechanism. For the sociology of institutions, the implication is that the coordinative mechanisms of the classical account operate at the level of observable interactional traces, a level at which participants with minimal cognition can take part and be measured on the same terms as human participants. For the study of human–AI collaboration, the implication is that the most consequential property of an automated teammate may not be its technical output but the form of its traces: whether it is stable enough to be re-encountered, addressable enough to be remembered, and predictable enough to be organized around.

## References

AIDev. 2026. "AIDev: Studying AI Coding Agents on GitHub." arXiv:2602.09185.

Anthony, Callen, Beth A. Bechky, and Anne-Laure Fayard. 2023. "'Collaborating' with AI: Taking a System View to Explore the Future of Work." *Organization Science* 34(5).

Ashenfelter, Orley. 1978. "Estimating the Effect of Training Programs on Earnings." *Review of Economics and Statistics* 60(1):47–57.

Ashery, Ariel Flint, Luca Maria Aiello, and Andrea Baronchelli. 2025. "Emergent Social Conventions and Collective Bias in LLM Populations." *Science Advances* 11:eadu9368.

Axelrod, Robert. 1984. *The Evolution of Cooperation*. New York: Basic Books.

Baker, Ted, and Reed E. Nelson. 2005. "Creating Something from Nothing: Resource Construction through Entrepreneurial Bricolage." *Administrative Science Quarterly* 50(3):329–366.

Barley, Stephen R. 1986. "Technology as an Occasion for Structuring: Evidence from Observations of CT Scanners and the Social Order of Radiology Departments." *Administrative Science Quarterly* 31(1):78–108.

Berger, Peter L., and Thomas Luckmann. 1966. *The Social Construction of Reality: A Treatise in the Sociology of Knowledge*. Garden City, NY: Doubleday.

Burt, Ronald S. 1992. *Structural Holes: The Social Structure of Competition*. Cambridge, MA: Harvard University Press.

Callaway, Brantly, and Pedro H. C. Sant'Anna. 2021. "Difference-in-Differences with Multiple Time Periods." *Journal of Econometrics* 225(2):200–230.

Cerulo, Karen A. 2009. "Nonhumans in Social Interaction." *Annual Review of Sociology* 35:531–552.

Collins, Randall. 2004. *Interaction Ritual Chains*. Princeton, NJ: Princeton University Press.

Dabbish, Laura, Colleen Stuart, Jason Tsay, and Jim Herbsleb. 2012. "Social Coding in GitHub: Transparency and Collaboration in an Open Software Repository." Pp. 1277–86 in *Proceedings of the ACM 2012 Conference on Computer Supported Cooperative Work*. New York: ACM.

DiMaggio, Paul J., and Walter W. Powell. 1983. "The Iron Cage Revisited: Institutional Isomorphism and Collective Rationality in Organizational Fields." *American Sociological Review* 48(2):147–160.

Eghbal, Nadia. 2020. *Working in Public: The Making and Maintenance of Open Source Software*. San Francisco: Stripe Press.

Espinosa, J. Alberto, Sandra A. Slaughter, Robert E. Kraut, and James D. Herbsleb. 2007. "Familiarity, Complexity, and Team Performance in Geographically Distributed Software Development." *Organization Science* 18(4):613–630.

Esposito, Elena. 2022. *Artificial Communication: How Algorithms Produce Social Intelligence*. Cambridge, MA: MIT Press.

Feldman, Martha S., and Brian T. Pentland. 2003. "Reconceptualizing Organizational Routines as a Source of Flexibility and Change." *Administrative Science Quarterly* 48(1):94–118.

Fine, Gary Alan. 1979. "Small Groups and Culture Creation: The Idioculture of Little League Baseball Teams." *American Sociological Review* 44(5):733–745.

Geiger, R. Stuart. 2011. "The Lives of Bots." Pp. 78–93 in *Critical Point of View: A Wikipedia Reader*, edited by G. Lovink and N. Tkacz. Amsterdam: Institute of Network Cultures.

Geiger, R. Stuart. 2014. "Bots, Bespoke Code and the Materiality of Software Platforms." *Information, Communication & Society* 17(3):342–356.

Geiger, R. Stuart, and Aaron Halfaker. 2017. "Operationalizing Conflict and Cooperation between Automated Software Agents in Wikipedia: A Replication and Expansion of 'Even Good Bots Fight.'" *Proceedings of the ACM on Human-Computer Interaction* 1(CSCW):49.

Giddens, Anthony. 1984. *The Constitution of Society: Outline of the Theory of Structuration*. Berkeley: University of California Press.

Goodman-Bacon, Andrew. 2021. "Difference-in-Differences with Variation in Treatment Timing." *Journal of Econometrics* 225(2):254–277.

Goyal, Agam, Olivia Pal, Hari Sundaram, Eshwar Chandrasekharan, and Koustuv Saha. 2026. "Social Simulacra in the Wild: AI Agent Communities on Moltbook." arXiv:2603.16128.

Halbwachs, Maurice. 1992 [1925]. *On Collective Memory*. Chicago: University of Chicago Press.

Halfaker, Aaron, R. Stuart Geiger, Jonathan T. Morgan, and John Riedl. 2013. "The Rise and Decline of an Open Collaboration System: How Wikipedia's Reaction to Popularity Is Causing Its Decline." *American Behavioral Scientist* 57(5):664–688.

Hallett, Tim, and Marc J. Ventresca. 2006. "Inhabited Institutions: Social Interactions and Organizational Forms in Gouldner's Patterns of Industrial Bureaucracy." *Theory and Society* 35(2):213–236.

Hassan, Ahmed E., et al. 2025. "The Rise of AI Teammates in Software Engineering (SE) 3.0: How Autonomous Coding Agents Are Reshaping Software Engineering." arXiv:2507.15003.

Hausman, Catherine, and David S. Rapson. 2018. "Regression Discontinuity in Time: Considerations for Empirical Applications." *Annual Review of Resource Economics* 10:533–552.

Heckathorn, Douglas D. 1990. "Collective Sanctions and Compliance Norms: A Formal Theory of Group-Mediated Social Control." *American Sociological Review* 55(3):366–384.

Heckman, James J., and Jeffrey A. Smith. 1999. "The Pre-programme Earnings Dip and the Determinants of Participation in a Social Programme: Implications for Simple Programme Evaluation Strategies." *Economic Journal* 109(457):313–348.

Holme, Petter, and Milena Tsvetkova. 2025. "Artificially Intelligent Agents in the Social and Behavioral Sciences: A History and Outlook." arXiv:2510.05743.

Howison, James, and Kevin Crowston. 2014. "Collaboration through Open Superposition: A Theory of the Open Source Way." *MIS Quarterly* 38(1):29–50.

Hutchins, Edwin. 1995. *Cognition in the Wild*. Cambridge, MA: MIT Press.

Imai, Kosuke, Luke Keele, and Dustin Tingley. 2010. "A General Approach to Causal Mediation Analysis." *Psychological Methods* 15(4):309–334.

Jepperson, Ronald L. 1991. "Institutions, Institutional Effects, and Institutionalism." Pp. 143–63 in *The New Institutionalism in Organizational Analysis*, edited by W. W. Powell and P. J. DiMaggio. Chicago: University of Chicago Press.

Kellogg, Katherine C., Melissa A. Valentine, and Angèle Christin. 2020. "Algorithms at Work: The New Contested Terrain of Control." *Academy of Management Annals* 14(1):366–410.

Knorr Cetina, Karin. 1997. "Sociality with Objects: Social Relations in Postsocial Knowledge Societies." *Theory, Culture & Society* 14(4):1–30.

Kollock, Peter. 1994. "The Emergence of Exchange Structures: An Experimental Study of Uncertainty, Commitment, and Trust." *American Journal of Sociology* 100(2):313–345.

Köster, Raphael, Dylan Hadfield-Menell, Richard Everett, Laura Weidinger, Gillian K. Hadfield, and Joel Z. Leibo. 2022. "Spurious Normativity Enhances Learning of Compliance and Enforcement in Artificial Agents." *Proceedings of the National Academy of Sciences* 119(3):e2106028118.

Latour, Bruno. 1992. "Where Are the Missing Masses? The Sociology of a Few Mundane Artifacts." Pp. 225–258 in *Shaping Technology/Building Society: Studies in Sociotechnical Change*, edited by W. E. Bijker and J. Law. Cambridge, MA: MIT Press.

Lebeuf, Carlene, Margaret-Anne Storey, and Alexey Zagalsky. 2018. "Software Bots." *IEEE Software* 35(1):18–23.

Lewis, Kyle. 2003. "Measuring Transactive Memory Systems in the Field: Scale Development and Validation." *Journal of Applied Psychology* 88(4):587–604.

March, James G. 1991. "Exploration and Exploitation in Organizational Learning." *Organization Science* 2(1):71–87.

Mead, George Herbert. 1934. *Mind, Self, and Society from the Standpoint of a Social Behaviorist*. Chicago: University of Chicago Press.

Meyer, John W., and Ronald L. Jepperson. 2000. "The 'Actors' of Modern Society: The Cultural Construction of Social Agency." *Sociological Theory* 18(1):100–120.

Meyer, John W., and Brian Rowan. 1977. "Institutionalized Organizations: Formal Structure as Myth and Ceremony." *American Journal of Sociology* 83(2):340–363.

Nass, Clifford, and Youngme Moon. 2000. "Machines and Mindlessness: Social Responses to Computers." *Journal of Social Issues* 56(1):81–103.

North, Douglass C. 1990. *Institutions, Institutional Change and Economic Performance*. Cambridge: Cambridge University Press.

Olick, Jeffrey K., and Joyce Robbins. 1998. "Social Memory Studies: From 'Collective Memory' to the Historical Sociology of Mnemonic Practices." *Annual Review of Sociology* 24:105–140.

O'Mahony, Siobhán, and Fabrizio Ferraro. 2007. "The Emergence of Governance in an Open Source Community." *Academy of Management Journal* 50(5):1079–1106.

O'Neill, Thomas, Nathan McNeese, Amy Barron, and Beau Schelble. 2022. "Human–Autonomy Teaming: A Review and Analysis of the Empirical Literature." *Human Factors* 64(5):904–938.

Orlikowski, Wanda J. 2000. "Using Technology and Constituting Structures: A Practice Lens for Studying Technology in Organizations." *Organization Science* 11(4):404–428.

Ostrom, Elinor. 1990. *Governing the Commons: The Evolution of Institutions for Collective Action*. Cambridge: Cambridge University Press.

Padgett, John F., and Walter W. Powell. 2012. *The Emergence of Organizations and Markets*. Princeton, NJ: Princeton University Press.

Page, Scott E. 2007. *The Difference: How the Power of Diversity Creates Better Groups, Firms, Schools, and Societies*. Princeton, NJ: Princeton University Press.

Powell, Walter W., and Jeannette A. Colyvas. 2008. "Microfoundations of Institutional Theory." Pp. 276–298 in *The SAGE Handbook of Organizational Institutionalism*, edited by R. Greenwood, C. Oliver, K. Sahlin, and R. Suddaby. London: SAGE.

Premack, David, and Guy Woodruff. 1978. "Does the Chimpanzee Have a Theory of Mind?" *Behavioral and Brain Sciences* 1(4):515–526.

Rahwan, Iyad, Manuel Cebrian, Nick Obradovich, Josh Bongard, Jean-François Bonnefon, Cynthia Breazeal, Jacob W. Crandall, et al. 2019. "Machine Behaviour." *Nature* 568(7753):477–86.

Reagans, Ray, Linda Argote, and Daria Brooks. 2005. "Individual Experience and Experience Working Together: Predicting Learning Rates from Knowing Who Knows What and Knowing How to Work Together." *Management Science* 51(6):869–881.

Rebatchi, Hocine, Tegawendé F. Bissyandé, and Naouel Moha. 2024. "Dependabot and Security Pull Requests: A Large Empirical Study." *Empirical Software Engineering* 29(5):128.

Roth, Jonathan. 2022. "Pretest with Caution: Event-Study Estimates after Testing for Parallel Trends." *American Economic Review: Insights* 4(3):305–322.

Seeber, Isabella, Eva Bittner, Robert O. Briggs, Triparna de Vreede, Gert-Jan de Vreede, Aaron Elkins, Ronald Maier, et al. 2020. "Machines as Teammates: A Research Agenda on AI in Team Collaboration." *Information & Management* 57(2):103174.

Sewell, William H., Jr. 1992. "A Theory of Structure: Duality, Agency, and Transformation." *American Journal of Sociology* 98(1):1–29.

Shirado, Hirokazu, and Nicholas A. Christakis. 2017. "Locally Noisy Autonomous Agents Improve Global Human Coordination in Network Experiments." *Nature* 545(7654):370–374.

Song, Fangchen, Ashish Agarwal, and Wen Wen. 2024. "The Impact of Generative AI on Collaborative Open-Source Software Development: Evidence from GitHub Copilot." arXiv:2410.02091.

Star, Susan Leigh. 1999. "The Ethnography of Infrastructure." *American Behavioral Scientist* 43(3):377–391.

Star, Susan Leigh, and Karen Ruhleder. 1996. "Steps Toward an Ecology of Infrastructure: Design and Access for Large Information Spaces." *Information Systems Research* 7(1):111–134.

Stinchcombe, Arthur L. 1965. "Social Structure and Organizations." Pp. 142–93 in *Handbook of Organizations*, edited by J. G. March. Chicago: Rand McNally.

Strachan, James W. A., Dalila Albergo, Giulia Borghini, Oriana Pansardi, Eugenio Scaliti, Saurabh Gupta, Krati Saxena, et al. 2024. "Testing Theory of Mind in Large Language Models and Humans." *Nature Human Behaviour* 8:1285–1295.

Sun, Liyang, and Sarah Abraham. 2021. "Estimating Dynamic Treatment Effects in Event Studies with Heterogeneous Treatment Effects." *Journal of Econometrics* 225(2):175–199.

Traeger, Margaret L., Sarah Strohkorb Sebo, Malte Jung, Brian Scassellati, and Nicholas A. Christakis. 2020. "Vulnerable Robots Positively Shape Human Conversational Dynamics in a Human–Robot Team." *Proceedings of the National Academy of Sciences* 117(12):6370–6375.

Tsvetkova, Milena, Ruth García-Gavilanes, Luciano Floridi, and Taha Yasseri. 2017. "Even Good Bots Fight: The Case of Wikipedia." *PLoS ONE* 12(2):e0171774.

Tsvetkova, Milena, Taha Yasseri, Niccolò Pescetelli, and Tobias Werner. 2024. "A New Sociology of Humans and Machines." *Nature Human Behaviour* 8:1994–2006.

Uzzi, Brian. 1997. "Social Structure and Competition in Interfirm Networks: The Paradox of Embeddedness." *Administrative Science Quarterly* 42(1):35–67.

Uzzi, Brian, and Jarrett Spiro. 2005. "Collaboration and Creativity: The Small World Problem." *American Journal of Sociology* 111(2):447–504.

VanderWeele, Tyler J. 2015. *Explanation in Causal Inference: Methods for Mediation and Interaction*. New York: Oxford University Press.

von Krogh, Georg, Sebastian Spaeth, and Karim R. Lakhani. 2003. "Community, Joining, and Specialization in Open Source Software Innovation: A Case Study." *Research Policy* 32(7):1217–1241.

Wegner, Daniel M. 1987. "Transactive Memory: A Contemporary Analysis of the Group Mind." Pp. 185–208 in *Theories of Group Behavior*, edited by B. Mullen and G. R. Goethals. New York: Springer-Verlag.

Wessel, Mairieli, Bruno Mendes de Souza, Igor Steinmacher, Igor S. Wiese, Ivanilton Polato, Ana Paula Chaves, and Marco A. Gerosa. 2018. "The Power of Bots: Characterizing and Understanding Bots in OSS Projects." *Proceedings of the ACM on Human-Computer Interaction* 2(CSCW):Article 182.

Wessel, Mairieli, and Igor Steinmacher. 2020. "What to Expect from Code Review Bots on GitHub? A Survey with OSS Maintainers." Pp. 457–462 in *Proceedings of the XXXIV Brazilian Symposium on Software Engineering*. New York: ACM.

Wessel, Mairieli, Igor Steinmacher, Igor Wiese, and Marco A. Gerosa. 2021. "Don't Disturb Me: Challenges of Interacting with Software Bots on Open Source Software Projects." *Proceedings of the ACM on Human-Computer Interaction* 5(CSCW2):1–21.

Wessel, Mairieli, Alexander Serebrenik, Igor Wiese, Igor Steinmacher, and Marco A. Gerosa. 2022. "Quality Gatekeepers: Investigating the Effects of Code Review Bots on Pull Request Activities." *Empirical Software Engineering* 27(5):108.

Yasseri, Taha, Robert Sumi, András Rung, András Kornai, and János Kertész. 2012. "Dynamics of Conflicts in Wikipedia." *PLoS ONE* 7(6):e38869.

Zhao, Eric Yanfei, Greg Fisher, Michael Lounsbury, and Danny Miller. 2017. "Optimal Distinctiveness: Broadening the Interface between Institutional Theory and Strategic Management." *Strategic Management Journal* 38(1):93–113.

Zucker, Lynne G. 1977. "The Role of Institutionalization in Cultural Persistence." *American Sociological Review* 42(5):726–43.